\documentclass[10pt,twocolumn,letterpaper]{article}

\usepackage{iccv}
\usepackage{booktabs} %

\usepackage[utf8]{inputenc}

\usepackage{times}
\usepackage{epsfig}
\usepackage{graphicx}
\usepackage{amsmath}
\usepackage{amssymb}
\usepackage{subcaption}
\usepackage{array}
\usepackage{multirow}
\usepackage{microtype}
\DeclareMathOperator*{\argmin}{argmin}

\usepackage{enumitem}
\setlist{nolistsep}

\usepackage{color}

\renewcommand{\etal}{{\em et~al.}\xspace}

\newcommand{\mlin}{\ensuremath{m_\mathsf{lin}}}
\newcommand{\clin}{\ensuremath{c_\mathsf{lin}}}
\newcommand{\mcyc}{\ensuremath{m_\mathsf{cyc}}}

\newcommand{\flinear}{\ensuremath{f_\mathsf{linear}}}
\newcommand{\fsin}{\ensuremath{f_\text{sin}}}
\newcommand{\fcyclic}{\ensuremath{f_\mathsf{cyclic}}}

\newcommand{\leaveout}[1]{}

\usepackage[pagebackref=true,breaklinks=true,letterpaper=true,colorlinks,bookmarks=false]{hyperref}

\iccvfinalcopy %

\def\iccvPaperID{5348} %

\ificcvfinal\fi

\begin{document}

\title{GeoStyle: Discovering Fashion Trends and Events}

\author{Utkarsh Mall$^{1}$ \\ {\tt\small utkarshm@cs.cornell.edu}\and
Kevin Matzen$^{2}$  \\ {\tt\small matzen@fb.com} \and
Bharath Hariharan$^{1}$ \\ {\tt\small bharathh@cs.cornell.edu} \and
Noah Snavely$^{1}$  \\ {\tt\small snavely@cs.cornell.edu} \and
Kavita Bala$^{1}$  \\ {\tt\small kb@cs.cornell.edu}
\end{tabular}
\\
\begin{tabular}[t]{c}%
$^{1}$Cornell University, $^{2}$Facebook
}

\maketitle
\ificcvfinal\thispagestyle{empty}\fi

\begin{abstract}
	Understanding fashion styles and trends is of great
	potential interest to retailers and consumers alike. 
	The photos people upload to social media are a historical and public data source of how people dress
	across the world and at different times. 
	While we now have tools to automatically recognize the clothing and style attributes of what people are wearing in these photographs, we lack the ability to analyze spatial and temporal trends in these attributes or make predictions about the future.
	In this paper we address this need by providing an automatic framework that analyzes large corpora of street
	imagery to (a) discover and forecast long-term trends of various fashion
	attributes as well as automatically discovered styles, and (b) identify spatio-temporally localized events that affect what
	people wear. 
	We show that our framework makes long term trend forecasts that are $>20\%$ more accurate than prior art, and identifies hundreds of socially meaningful events that impact fashion across the globe. The supplementary material can be found at \href{https://geostyle.cs.cornell.edu/static/pdf/supplementary.pdf}{https://geostyle.cs.cornell.edu/static/pdf/supplementary.pdf}
\end{abstract}

\maketitle

\section{Introduction}
	
Each day, we collectively upload to social media platforms billions of photographs 
that capture a wide range of human life and activities around the world.
At the same time, object detection, semantic segmentation, and visual search
are seeing rapid advances~\cite{he-17} and are being deployed at scale~\cite{mahajan2018exploring}.
With large-scale recognition available as a fundamental tool in our vision
toolbox, it is now possible to ask questions about how people dress, eat,
and group across the world and over time. In this paper we focus on how people dress.
In particular, we ask: can we \emph{detect and predict} fashion trends and
styles over space and time?

We answer these questions by designing
an automated method to characterize and predict seasonal and year-over-year fashion trends, 
detect social events (e.g., festivals or sporting events) that impact how people dress, and identify social-event-specific
style elements that epitomize these events.
Our approach uses existing recognition algorithms to identify a
coarse set of fashion attributes in a large corpus of images. We then
fit interpretable parametric models of long-term temporal trends to these fashion
attributes.
 These models capture both seasonal cycles as well as changes in popularity over time.
These models not only help in understanding existing trends, but can also make up to 20\% more accurate, temporally fine-grained forecasts across long time scales compared to
prior methods~\cite{AlHalah-17}.  For example, we find that year-on-year more people are wearing black, but that they tend to do so more in the winter than in the summer.

Our framework not only models long-term trends, but also identifies sudden, short-term changes in popularity that buck these trends. 
We find that these outliers often correspond
to festivals, sporting events, or other large social gatherings. 
We provide a methodology to 
automatically \emph{discover} the events underlying such outliers by looking
at associated image tags and captions, thus tying visual analysis to text-based discovery.
We find that our framework finds understandable reasons for \emph{all} of the most salient events it discovers, and in so doing surfaces intriguing social events around the world that were unknown to the authors.
For example, it discovers an unusual increase in the color yellow in
Bangkok in early December, and associates it with the words
``father'', ``day'', ``king'', ``live'', and ``dad''. 
This corresponds to the king's birthday, celebrated as Father's Day in Thailand by wearing yellow~\cite{fathersday}.
Our framework similarly surfaces events in Ukraine (Vyshyvanka Day), Indonesia (Batik Day), and Japan (Golden Week). 
Figure \ref{fig:fig1} shows more of the worldwide events discovered by our framework and the clothes that people wear during those events.

\begin{figure*}
\centering
\includegraphics[width=\linewidth]{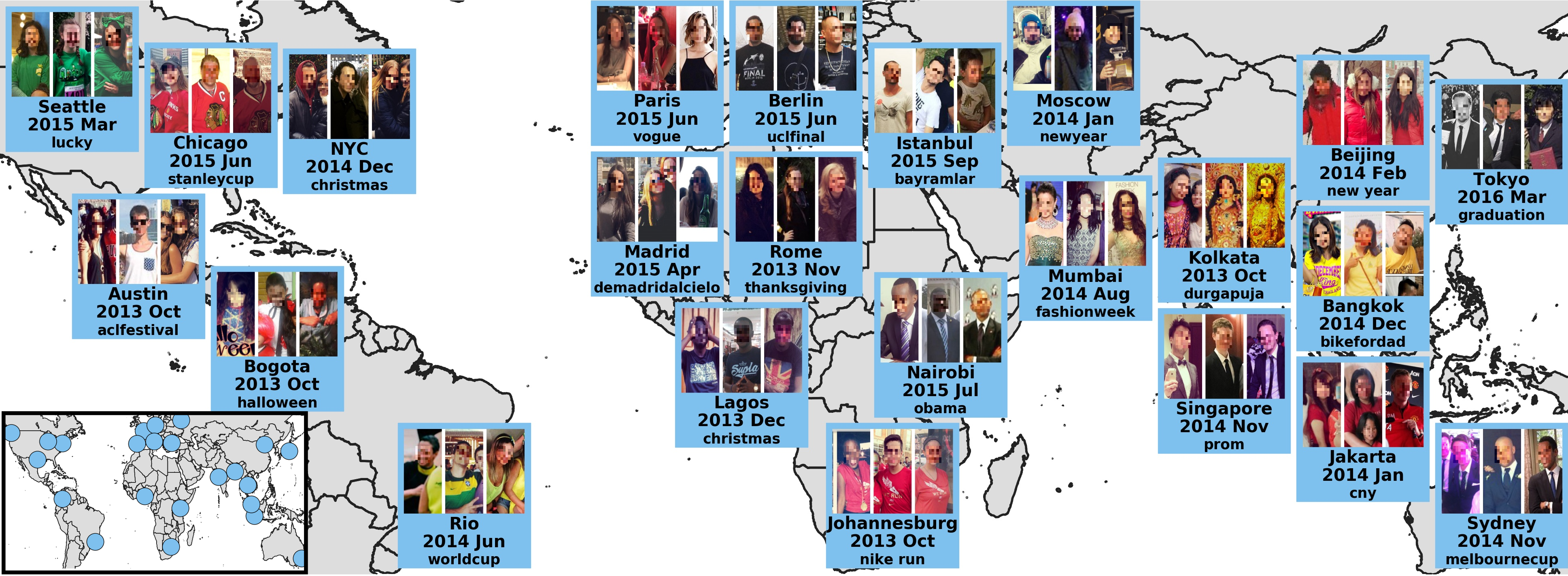}
	\caption{Major events discovered by our framework. 
	For each event, the figure shows the clothing that people typically wear for that event, along with the city, one of the months of occurrence, and the most descriptive word extracted using the images captions. The inset image shows more precise locations of these cities.}
\label{fig:fig1}
\end{figure*}

We further show that we can predict trends and events not just at the level of individual fashion attributes (such as ``wearing yellow''), but also at the level of \emph{styles} consisting of recurring visual ensembles. These styles are identified by clustering photographs in feature space to reveal \emph{style clusters}: clusters of people dressed in a similar style.
Our forecasts of the future popularity of \emph{styles} are just as accurate as our predictions of individual attributes.
Further, we can run the same event detection framework described above on style trends, 
allowing us to not only automatically detect social events, but also associate each event with its own distinctive style; a \emph{stylistic signature} for each event.

Our contributions, highlighted in Figure \ref{fig:pipeline}, include:
\begin{itemize}
	\item We present an automated framework for analyzing the temporal behavior of fashion elements across the globe. Our framework models and forecasts long-term trends and seasonal behaviors. It also automatically identifies short-term spikes caused by events like festivals and sporting events.
	\item Our framework automatically discovers the reasons behind these events by leveraging textual descriptions and captions. 
 \item We connect events with signature \emph{styles} by performing this analysis on automatically discovered style clusters.

\end{itemize}

\section{Related work}

\noindent \textbf{Visual understanding of clothing.}
There has been extensive recent work in computer vision on characterizing clothing.
Some of this work recognizes attributes of people's clothing, such as whether a shirt has short or long
sleeves~\cite{Chen-12,Bourdev-11,Bossard-12,Zhang-14, Liu-16, matzen-17}. 
Other work goes beyond coarse image-level labels and attempts to segment different clothing items in images~\cite{Yamaguchi-12,Yamaguchi-13,Yang-14}.
Product identification
is an ``instance-level classification'' task used for detecting specific
clothing products in photos~\cite{Di-13,Vittayakorn-15,Kiapour-15}.  
Finally, there is also prior work on classifying the ``style'': the ensemble of clothing a person is wearing, e.g., ``hipster'', ``goth'' etc.~\cite{Kiapour-14}.
In some cases, these labels might be unknown and require discovery~\cite{matzen-17,Hsiao-17}, often by leveraging embeddings of images learnt by attribute recognition systems.

Our work borrows from the attribute and style literature.  We make use
of several human-annotated attributes on a small dataset to
form an embedding space for the exploration of a much larger set of images.
We use the embedding space to label attributes and styles over a vast internet-scale dataset.
However, our goal is not the labeling itself, but the \emph{discovery} of interesting geo-temporal trends and their associated 
styles.

\smallskip \noindent \textbf{Visual discovery.} 
Although less common, there has been some prior research into using visual analysis to identify trends.
Early work used low-level image features or mined visually distinctive patches~\cite{Doersch-12, Singh-12, Doersch-13} to predict geo-spatial properties such as perceived safety of cities~\cite{Arietta-14,Naik-14,Ordonez-14}, or ecological properties such as snow or cloud cover~\cite{Zhang-12,Wang-13,Murdock-15}. 
Advances in visual recognition has enabled more sophisticated analysis, such as the analysis of demographics by recognizing the make and model of cars in Street View~\cite{Gebru-17}.
However, while this work is exciting, the focus has been on using vision to predict known geo-spatial trends rather than discover new ones.
The notion of using visual recognition to power \emph{discovery} and \emph{prediction of the future} is under-explored.
Some initial research in this regard has focused on faces~\cite{Islam-15, Salem-16, Ginosar-15} and on human activities in a healthcare setting~\cite{luo2018computer}.
However, this prior work has mostly focused on descriptive analytics and manual exploration of the data to discover interesting trends.
By contrast, we propose an automated, quantitative framework for both long-term forecasting and discovery. 
While our work focuses on the fashion domain, our ideas might be adapted to other applications as well.

\smallskip \noindent \textbf{Trend analysis in fashion.} 
Trend analysis has also been applied to the fashion domain, the focus of our work.
Often, prior work has considered small datasets such as catwalk images from NYC fashion shows~\cite{Hidayati-14}.
Where larger datasets have been analyzed, interesting trends have been discovered, such as a sudden increase in popularity for
``heels'' in Manila~\cite{Simo-Serra-15} or seasonal trends related to styles such as ``floral", ``pastel", and ``neon"~\cite{Vittayakorn-15}. 
Matzen~\etal\ ~\cite{matzen-17} significantly expand the scope of such trend discovery by leveraging publicly available images uploaded to social media. We build upon the StreetStyle dataset in this work.
However, the analysis of the spatial and temporal trends in these papers is often descriptive, and their use for discovery requires significant manual exploration.
The first problem is partly addressed by Al-Halah \etal~\cite{AlHalah-17}, who attempt to make quantitative forecasts of fashion trends, but whose temporal models are limited in their expressivity, forcing them to make very coarse yearly predictions for just one year in advance.
In contrast, we propose an expressive parametric model for trends that makes much higher quality, fine-grained weekly predictions for as much as 6 months in advance.
In addition, we propose a framework that automates discovery by automatically surfacing interesting outlier events for analysis.

\section{Method}\label{sec:method}
\begin{figure}
\centering
\includegraphics[width=\linewidth]{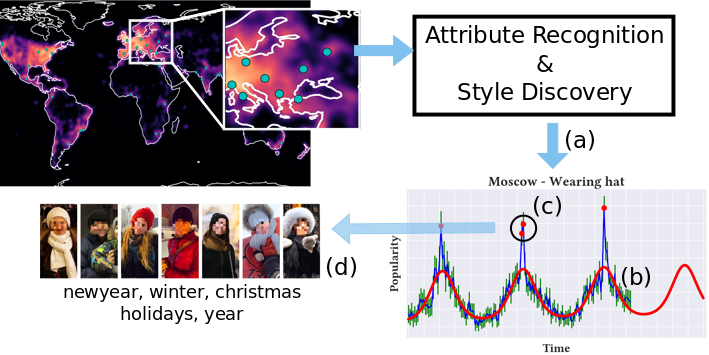}
	\caption{Approach overview. 	(a) Attribute recognition and style discovery~\cite{matzen-17} on internet images from multiple cities gives us temporal trends. (b) We fit interpretable parametric models to these trends to characterize and \emph{forecast} (red curve is the fitted trend used to forecast). (c) Deviations from parametric models are identified as events (red points). (d) We identify text and styles specific to each event.}
\label{fig:pipeline}
\end{figure}

Our overall pipeline is shown in Figure~\ref{fig:pipeline}.
We first describe our dataset and fashion attribute recognition pipeline, which we adapt from StreetStyle~\cite{matzen-17} and then describe our trend analysis and event detection pipeline.

\subsection{Background: dataset and attribute recognition}
\label{ssec:data}
Our dataset uses photos from two social media websites, Instagram and Flickr.
In particular, we start with the Instagram-based StreetStyle dataset of Matzen~\etal~\cite{matzen-17} and extend
it to include photos from the Flickr 100M dataset~\cite{Thomee-16}.
The same pre-processing applied to StreetStyle is also applied to
Flickr 100M, including categorization of photos into 44 major world cities across 6 continents,
person body and face detection, and canonical cropping.  Please refer to~\cite{matzen-17} for details.
In total, our dataset includes 7.7 million images of people from around the world.

\label{sec:learning_attributes}
Matzen~\etal\ also collect clothing attribute annotations on a 27k subset of the StreetStyle dataset~\cite{matzen-17}.
As in their work, we use these annotations to train a multi-task CNN (GoogLeNet~\cite{Szegedy-15}) where separate heads predict separate attributes, e.g., one head
may predict ``long-sleeves'' whereas another may predict ``mostly yellow''. 
This training also has the effect of automatically producing an embedding of images in the penultimate layer of the network that places similar clothing attributes and combinations of these attributes, henceforth refered to as ``styles'', into the same
region of the embedding vector space.

We take these attribute classifiers and apply them to the full unlabeled
set of 7.7M of people images.
We produce a temporal trend for each attribute in each city by computing, for each week, the mean probability of an attribute across all photos from that week and city.
Per-image probabilities are derived from the CNN prediction scores after calibration via isotonic regression on a validation set~\cite{matzen-17}.

\subsection{Characterizing trends}
\label{ssec:trend_prediction}
\begin{figure}
\centering
\includegraphics[width=\linewidth]{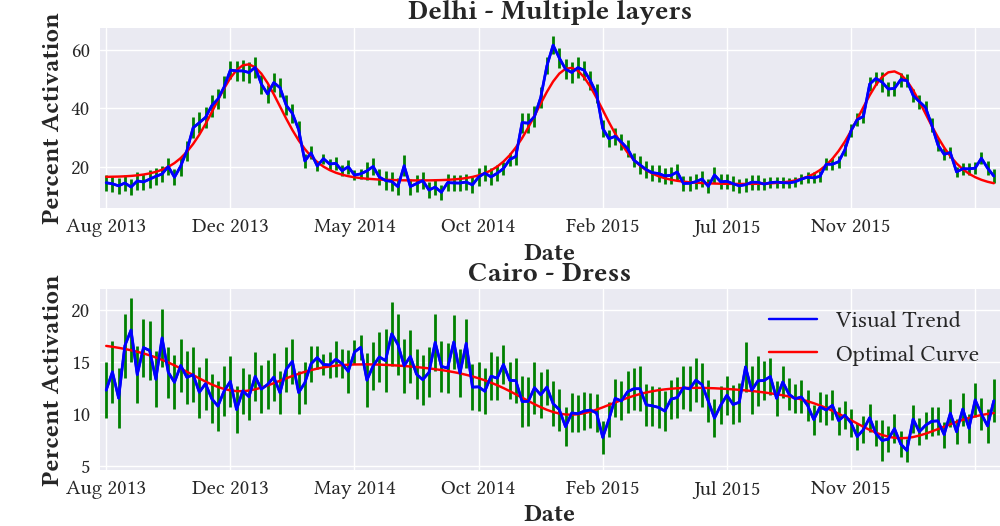}
	\caption{Two examples of observed trends. As can be seen, trends often have seasonal variations, but periodic trends are not necessarily sinusoidal. Trends can also involve a linear component (e.g., the decrease in the incidence of Dresses in Cairo over time). The green bars indicate the 95\% confidence interval for each week.}
\label{fig:unfitted_trend}
\end{figure}
Given each weekly clothing attribute trend in each city, we wish to (a) characterize this trend in a human-interpretable manner, and (b) make accurate forecasts about where the trend is headed in the future.

Figure~\ref{fig:unfitted_trend} shows two examples of attribute trends over time.
We observe several behaviors in these examples. 
First, there are both coarse-level trends extending over months or years (e.g., the seasonal cycles in the wearing of multiple layers in Delhi) as well as fine-scale spikes that occur over days or weeks (e.g., the spike in December 2014).
Second, the coarse trend often has a strong periodic component usually governed by different seasons.
Third, instead of even sinusoidal upswings and downswings, the periodic trend often consists of upward (Figure \ref{fig:unfitted_trend} top) or downward (Figure \ref{fig:unfitted_trend} bottom) \emph{surges} in popularity.
Fourth, in some cases this periodic trend is superimposed on a more gradual increase or decrease in popularity, as in Figure \ref{fig:unfitted_trend} (bottom).
\begin{figure}
\centering
\includegraphics[width=\linewidth]{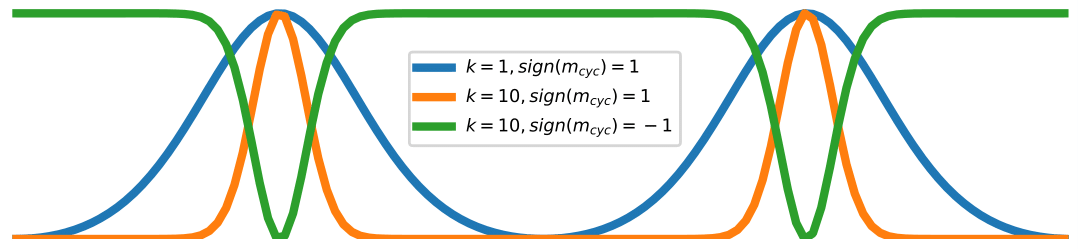}
	\caption{We use a function of the form $\mcyc e^{k\sin(\omega x + \phi)-k}$ as our cyclical component because of its ability to model seasonal spikes. This plot shows this function for three values of $k$ and $\mcyc$. For ease of comparison, all three functions have been centered and rescaled to the same dynamic range.}
\label{fig:exp}
\end{figure}

We seek to identify both the coarse, slow-changing trends that are governed by seasonal cycles or slow changes in popularity, as well as the fine-grained spikes that might arise from events such as festivals (Christmas, Chinese New Year) or sporting events (FIFA World Cup).
The former might tell us how people in a particular place dress in different seasons, while the latter might reveal important social events with many participants.
We first fit a parametric model to capture the slow-changing trends (this section), and then identify potential events as large departures from the predicted trends (Section~\ref{ssec:events}).

We model slow-changing trends using a parametric model $f_\theta(t)$, which is a convex combination of two components: a linear component and a cyclical component:
\begin{equation}\label{eq:curvefit}
  f_\theta(t) = (1-r) \cdot L(t) + r \cdot C(t)
\end{equation}
where the parameter $r\in[0,1]$ defines the contribution of each component.
The linear component, $L(t)$ is characterized by slope $\mlin$ and intercept $\clin$:
\begin{equation}
  L(t) = \mlin t+\clin
\end{equation}
A standard choice for the cyclical component would be a sinusoid. However, we want to capture upward and downward surges, so we instead use a more expressive cyclical component of the form:
\begin{equation}
  C(t) = \mcyc e^{k\sin(\omega t + \phi)-k}.
\end{equation}
When $k$ is close to $0$, this function behaves like a (shifted) sinusoid, but for higher values of $k$, it has more peaky cycles (Figure~\ref{fig:exp}). 
$\omega$ and $\phi$ denote period and phase respectively.

The full set of parameters in this parametric model is $\theta = \{r,\mcyc,k,\omega,\phi,\mlin,\clin\}$.
Table~\ref{tab:intuitive} provides intuitive descriptions of these parameters.
Because each parameter is interpretable, our model allows us to not just make predictions about the future but also to \emph{discover} interesting trends and analyze them, as we show in Section~\ref{ssec:trend_prediction_res}.
\begin{table}
\resizebox{0.99\linewidth}{!}{
\begin{tabular}{ll}
\toprule
Parameter & Intuitive meaning \\
\midrule
$r$ & Trade-off between linear and cyclic trend \\
$\clin$ & Long term bias \\
$\mlin$ & Rate of long-term increase/decrease in popularity \\
$\mcyc$ & Amplitude and sign (upwards/downwards) of cyclical spikes \\
$k$ & Spikiness of cyclical spikes \\
$\omega$ & Frequency of cyclical spikes\\
$\phi$ & Phase of cyclical spikes\\
\bottomrule
\end{tabular}%
}
\caption{Intuitive descriptions of all parameters}
\label{tab:intuitive}
\end{table}

We fit the parameters $\theta$ of the above model to the weekly trend of each attribute for each city by solving the following non-linear least-squares problem:
\begin{equation}\label{eq:curvefitloss}
  \theta^{*} = \argmin_\theta \sum_t \left(\frac{f_\theta(t)-T(t)}{\sigma(t)}\right)^2
\end{equation}
where $T(t)$ represents the observed average probability of the attribute for week $t$ in the particular city/continent/world and $\sigma(t)$ measures the uncertainty of the measurement (reciprocal of the binomial confidence). 
We minimize Equation~\eqref{eq:curvefitloss} using the Trust Region Reflective algorithm~\cite{lukvsan-93}.
To prevent overfitting we set an upper bound for $\omega$ to keep seasonal variation close to annual variation.
We set it to $\frac{2\pi\times2}{52}$, allowing for a maximum of two complete sinusoidal cycles over a year. 
We chose 52 because we measure time in weeks.

\subsection{Discovering events}\label{ssec:events}
Given a fitted model, we now describe how we identify more fine-grained structure in each attribute trend, and correlate these structures with potentially important social gatherings.
In particular, we are interested in sharp spikes in popularity of particular kinds of clothing, which often are due to an event.
For example, people might wear a particular jersey to support their local team on game night, or wear green on St.\ Patrick's Day.

To discover such events, we start by identifying weeks with large, positive deviations from the fitted model, or \emph{outliers}, using a binomial hypothesis test.
The set of images in week $t$ are considered as a set of trials, with those images classified as positive for the attribute constituting ``successes'' and others failures.
The null hypothesis is that the probability of a success is given by the fitted parametric model, $f_\theta^*(t)$.
Because we are interested in \emph{positive} deviations from this expectation, we use a one-tailed hypothesis test, where the alternative hypothesis is that the true probability of success is greater than this expectation.
We identify outliers as weeks with $p$-value $<0.05$.
We use the reciprocal of the $p$-value, denoted by $s$, as a measure of outlier \emph{saliency}.

We then connect the outliers discovered to the social event that caused them, if any.
To do so, we note that some of these events might be repeating, annual affairs (such as festivals), while others might be one-off events (e.g. FIFA World Cup).
We therefore formalize an \emph{event} as a \emph{group of outliers} that are either localized on a few weeks (\textbf{one-off events}) or are separated by a period of approximately a year (\textbf{annual events}, like festivals on a solar or a lunisolar calendar~\cite{lunisolar}).

To determine if our detected outliers fit some event, we need a way to \emph{score} candidate events.
If we have a sequence of outliers $g = \{t_1, \ldots, t_k\}$ for a particular trend in a specific city, how do we say if this group of outliers is likely to be an actual event?
There are two main considerations in this determination.
First, the outliers involved in the event must be \emph{salient}, that is, they should correspond to significant departures from the background trend. 
Second, they should have the \emph{temporal signature} described above: the outliers involved should either be localized in time, or separated by approximately a year.

We formalize this intuition by defining a cost function $C(g)$ for each group of outliers $g = \{t_1,\ldots, t_k\}$ such that a smaller cost indicates a higher likelihood of $g$ being an event.
$C(g)$ is a product of two terms: a cost incentivizing the use of salient outliers (we use the reciprocal of the average saliency $\bar{s}$ of the outliers involved), and a cost $C_T(g)$ measuring the deviation from the ideal temporal signature:
\begin{equation}
C(g) = \frac{C_T(g)} {\bar{s}}
\end{equation}
$C_T(g)$ considers consecutive outliers in $g$ and assigns a low cost if these consecutive outliers occur very close to each other in time, or are very close to following an annual cycle.
If consecutive events are neither proximal (they are more than $\Delta_{\max}$ weeks apart) nor part of an annual or multi-year cycle (they miss the cycle by more than $d_{\max}$ weeks), the cost is set to infinity.
Concretely, we define $C_T$ as follows:
\begin{align}
C_T (g) &= \frac{\sum_{i=1}^{|g|-1} C_p(t_{i+1}-t_i)}{|g|-1} 
\end{align}
\begin{align}
C_p(\Delta) =\left\{\begin{array}{lr} \frac{\Delta + c}{\Delta_{\max} + c} & \textrm{ if } \Delta<\Delta_{\max} \\ 
				\multirow{3}{*}{$\frac{d(\Delta)+b}{d_{\max}+b}$} & \\
				   &  \textrm{ if } \Delta\geq T-d_{\max}\\
				   & \textrm{ and } d(\Delta) <d_{\max}\\
				\infty  & \textrm{ otherwise. }\end{array}\right.
\end{align}
Here, $|g|$ denotes the cardinality of outlier group $g$. $\Delta$ is the time difference between consecutive outliers, $T$ is the length of a year, and $d(\Delta)$ measures how far $\Delta$ is from an annual cycle. In particular, $d(\Delta) =  \min(\Delta \mod T,   -\Delta \mod T)$. 
$c = 18,b=15, \Delta_{\max}=2$ and $d_{\max}=5$ are constants. The setting of these is explained in the supplementary.
When $g$ contains a single event, $C_T(g)$ is defined to be 1. 

$C(g)$ gives us a way of scoring candidate events, but we still need to come up with a set of candidates in the first place from the discovered outliers.
There may be multiple events in a city over time (e.g., Christmas and Chinese New Year), and we need to separate these events.
We consider this as a \emph{grouping} problem: given a set of outliers occuring at times $t_1, \ldots, t_n$ in the trend of a particular attribute in a particular city, we want to \emph{partition} the set into groups.
Each group is then a candidate event.
We define the cost of a partition $P = \{g_1, \ldots, g_k\}$ as the average cost $C(g_i)$ of each group $g_i$ in the partition, and choose the partition that minimizes this cost:
\begin{equation}
P^* = \argmin_P \frac{\sum_i C(g_i)}{|P|} 
\end{equation}
This is a combinatorial optimization problem. However, we find that there are very few outliers for each trend, so this problem can be solved optimally using simple enumeration.

Running this optimization problem for each trend gives us a set of events, each corresponding to a group of outliers.
Each event is then associated with a cost $C(g)$.
We define the reciprocal of this cost as the \emph{saliency of the event}, and we rank the events in decreasing order of their saliency.

\medskip\noindent\textbf{Mining underlying causes for events.}
To derive explanations for each event, we analyze image captions that accompany the image dataset.
We consider images from the relevant location classified as positive for the relevant attribute across the year, and split them into two subsets: those appearing within the event weeks, and those at other times.
Words appearing in captions of the former but not the latter may indicate why the attribute is more popular specifically in that week.
To find these words, we do a TF-IDF \cite{Sparck-72} sorting, considering the captions of the first set as positive documents and the captions of the second set as negatives.
Images can contribute to a term at most once in term frequencies.
We perform this analysis using the English language captions.

\subsection{Style trend analysis}\label{ssec:style}

We also wish to identify trends not just in single attributes, but also in combinations of attributes that correspond to looks or styles.
However, the number of possible attribute combinations grows exponentially with the number of attributes considered, and most attribute combinations are uninteresting because of their rarity: e.g., pink, short-sleeved, suits.
Instead, we want to focus on the limited set of attribute combinations that are actually prevalent in the data.
To do so, we follow the work of Matzen~\etal~\cite{matzen-17} to discover \emph{style clusters}: popular combinations of attributes. 
Style clusters are identified using a Gaussian mixture model
to cluster images in the feature space learned by the CNN.
To ensure coverage of all trends while also maintaining sufficient data for each style cluster, we separately find a small number of style clusters in each city.
In general, we find that the style clusters we discover correspond to intuitive notions of style.
As with individual attributes, our trend analysis on these clusters tells us not only which styles are coming into or going out of fashion, but also associates styles with major social events (Section \ref{ssec:style_res}).

\section{Results}\label{sec:results}
We now evaluate our ability to discover and predict style events and trends. In addition, we visualize discovered trends, events, and styles.

\subsection{Trend prediction and analysis}\label{ssec:trend_prediction_res}
We first evaluate our parametric temporal model (Eq.~\ref{eq:curvefit}) based on its ability to make out-of-sample predictions about the future (in-sample predictions are provided in the supplementary).
We compare to models proposed by Al-Halah~\etal~\cite{AlHalah-17}, the most relevant prior work.
We also compare to four ablations of our model: (a) Linear: $\flinear(t) = \mlin t+c$, (b) Sinusoidal fit: $\fsin(t) = \sin(\omega t + \phi)$, (c) Cyclic fit: $\fcyclic(t) = \mcyc e^{k\sin(\omega t + \phi) - k}$ and (d) a linear combination of $\flinear$ and $\fsin$.
We use the same metrics as Al-Halah~\etal~\cite{AlHalah-17}, namely, MAE and MAPE.
The latter looks at the average absolute error relative to the true trend $T(t)$, expressed as a percentage.
However, while Al-Halah~\etal only evaluate prediction accuracy in the extreme short term (the very next data point), we consider prediction accuracy both in the \emph{short term} (next data point, or next week) as well as the long term (next 26 data points, or next 6 months). Note that even though Al-Halah~\etal~only evaluate predictions over the next data-point, that data point corresponds to a full year. Hence they are predicting trends farther in the future, but their prediction is relatively coarser. We also show the results of our prediction for more than one year in supplementary.

We find that our parametric model is significantly better than all baselines at both long-term and short-term predictions (see Table \ref{tab:prediction}).
Furthermore, the gap between our model and the best method found by Al-Halah \etal (exponential smoothing) \emph{increases} when we move to making long-term predictions.
We also observe that our model's out-of-sample performance actually matches in-sample performance (shown in supplementary) very well, indicating strong generalization.
This shows that our model generalizes better and can extrapolate significantly further into the future.

Interestingly, our model is also significantly better than the autoregressive baselines.
These baselines predict a data point as a weighted linear combination of the previous $k$ data points, where the weights are learned from data and $k$ is cross-validated.
Thus, these models have many more parameters than our model (up to 12$\times$ more). 
The fact that our model still performs better suggests that choosing the right parametric form is more important than merely the size or capacity of the model.

\begin{table}
\centering
      \textbf{Attribute-based trends}
      \begin{tabular}{@{}>{\arraybackslash}l >{\arraybackslash}r >{\arraybackslash}r >{\arraybackslash}r >{\arraybackslash}r@{}} 
        \toprule
        \textbf{Model} & \multicolumn{2}{c}{\textbf{Next week}} & \multicolumn{2}{c}{\textbf{Next 26 weeks}} \\
         & \multicolumn{1}{c}{MAE} & \multicolumn{1}{c}{MAPE} & \multicolumn{1}{c}{MAE} & \multicolumn{1}{c}{MAPE} \\
        \midrule
        mean & 0.0209 & 19.05 & 0.0292 & 25.79\\
        last & 0.0153 & 15.56 & 0.0226 & 21.04\\
        \midrule
        AR & 0.0147 & 14.18 & 0.0207 & 20.27\\
        VAR & 0.0146 & 16.16 & 0.0162 & 18.92\\
        ES & 0.0152 & 14.92 & 0.0231 & 20.59\\
        \bottomrule
        linear & 0.0276 & 18.35 & 0.0365 & 24.40\\
        sinusoid & 0.0141 & 13.22 & 0.0163 & 16.09\\
        sin+lin & 0.0140 & 13.17 & 0.0169 & 16.87\\
        cyclic & 0.0129 & 12.63 & 0.0165 & 16.64\\
        \bottomrule
        Ours & \textbf{0.0119} & \textbf{12.13} & \textbf{0.0145} & \textbf{15.73}\\
        \bottomrule
      \end{tabular}\\
      \vspace{1em}
      \textbf{Style-based trends}
      \begin{tabular}{@{}>{\arraybackslash}l >{\arraybackslash}r >{\arraybackslash}r >{\arraybackslash}l >{\arraybackslash}r >{\arraybackslash}r@{}} 
        \toprule
        \textbf{Model} & \multicolumn{2}{c}{\textbf{Next 26 weeks}}
          & \textbf{Model} & \multicolumn{2}{c}{\textbf{Next 26 weeks}} \\
          & \multicolumn{1}{c}{MAE} & \multicolumn{1}{c}{MAPE}
          & & \multicolumn{1}{c}{MAE} & \multicolumn{1}{c}{MAPE} \\
        \cmidrule(lr){1-3}\cmidrule(lr){4-6}
        mean & 0.0101 & 31.82 & linear & 0.0135 & 36.05\\
        last & 0.0145 & 44.57 & sinusoid & 0.0083 & 23.23\\
        \cmidrule(lr){1-3}
        AR & 0.0090 & 37.89 & sin+lin & 0.0081 & 23.04\\
        VAR & 0.0120 & 27.97 & cyclic & 0.0085 & 24.16 \\
        \cmidrule(lr){4-6}
        ES & 0.0143 & 43.96 & Ours & \textbf{0.0077} & \textbf{21.78}\\
        \bottomrule
      \end{tabular}
	\caption{Comparison of our prediction model against other models from \cite{AlHalah-17}. Mean and Last are naive methods that predict the mean and last of the known time series as the next prediction respectively. AR (autoregression) and VAR (vector-autoregression) are autoregressive methods. ES is exponential smoothing. Lower values are better.}
  \label{tab:prediction}
\end{table}

\medskip\noindent\textbf{Interpretability:}
Our model fitting characterizes each attribute trend in terms of a few interpretable parameters, shown in Table~\ref{tab:intuitive}, which can be used in a straightforward manner to reveal insights.
For example, $\phi$ describes the phase of the cyclical trend.
If we look at cities where there is a positive spike in people wearing multiple layers in the winter, then the peaks should occur in winter months, and cities in the northern and southern hemisphere should be exactly out of phase.
Figure~\ref{fig:phase} shows the difference in phase $\phi$ for the \emph{multiple-layered} clothing attribute between each pair of cities.
We find that cities indeed cluster together based on their hemisphere, with cities in the same hemisphere closer to each other in phase.
Interestingly, cities closer to the equator seem to be half-way between the two hemispheres and form their own cluster.

As another example, $k$ represents the ``spikiness'' of the cyclical trend: a high $k$ corresponds to a very short-duration increase/decrease in popularity. 
We can search for attribute trends that show the spikiest (i.e., highest $k$) annual positive spikes.
These turn out to be \emph{wearing-scarves} in Bangkok and \emph{clothing-category-dress} in Moscow.
This might reveal the fact that Bangkok has a very short winter when people wear scarves, while Moscow has a short summer where people wear dresses.

\subsection{Event discovery}\label{ssec:event_res}
After fitting our parametric trend model, we discover events using the method discussed in Section \ref{ssec:events}. 
Our event discovery pipeline detected hundreds of events, detailed in the supplementary.
Table~\ref{tab:events} shows the five most salient events along with the corresponding words associated with the event and a set of corresponding images.
All five correspond to significant social gatherings that some or all of the authors were unaware of \emph{a priori} :
\begin{enumerate}
\item \textbf{Father's Day in Bangkok} is celebrated on the King's birthday, and people wear yellow to honor the king.
\item \textbf{FreakNight in Seattle} is a dance music event held on or around Halloween. The prevalance of sleeveless clothes is an outlier driven by this event given cool weather at this time of the year.
\item \textbf{Songkran in Bangkok} is a festival celebrated in April on the Thai New Year and involves people playing with water in warm weather.
\item The Western Conference Finals of the \textbf{Stanley Cup 2014 in Chicago} involved the Chicago Blackhawks and the Los Angeles Kings. People wore their home team's jerseys.
\item The \textbf{FIFA World Cup} was held in Brazil in 2014 and featured a prevalence of yellow jerseys in support of Brazil.
\end{enumerate}
Note that events such as Father's Day were further correctly identified as annual events. 
\begin{figure}
\centering
\includegraphics[width=0.6\linewidth]{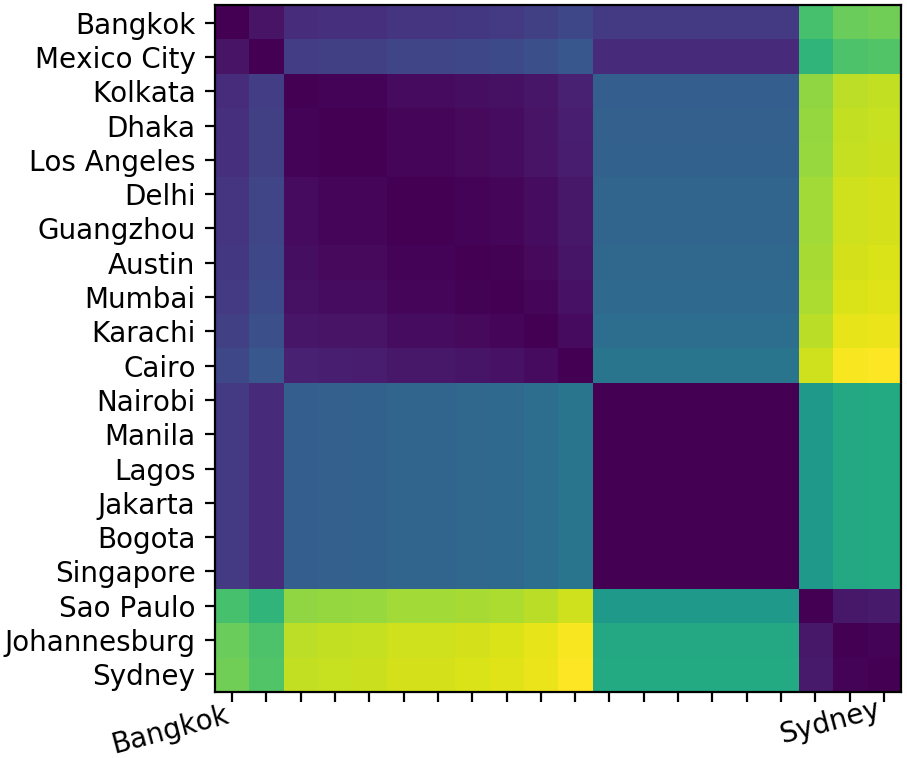}
	\caption{Phase difference for the \emph{multiple-layered} attribute between 20 cities, using estimated phase parameter $\phi$.}
\label{fig:phase}
\end{figure}

\begin{table*}
\centering
\begin{tabular}{@{} lccccc @{}} 
  \toprule
  \textbf{Images} &
  \includegraphics[width=0.15\linewidth]{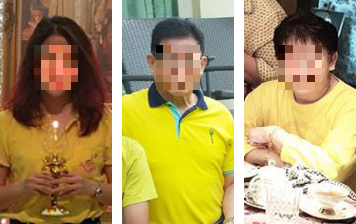} & 
  \includegraphics[width=0.15\linewidth]{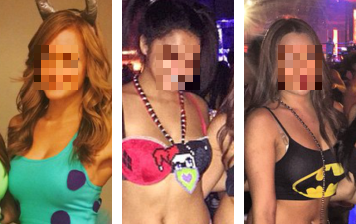} &
  \includegraphics[width=0.15\linewidth]{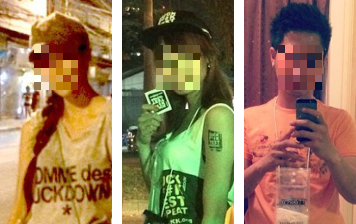} &
  \includegraphics[width=0.15\linewidth]{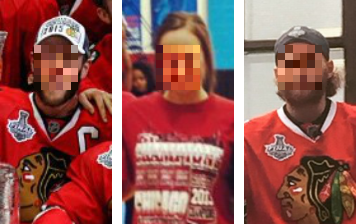} &
  \includegraphics[width=0.15\linewidth]{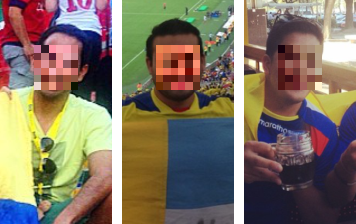} \\
  &
  \includegraphics[width=0.15\linewidth]{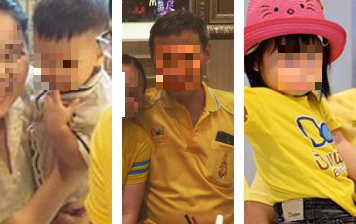} & 
  \includegraphics[width=0.15\linewidth]{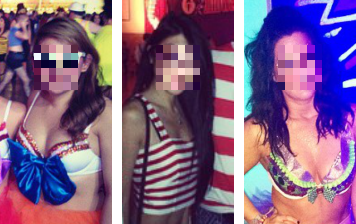} &
  \includegraphics[width=0.15\linewidth]{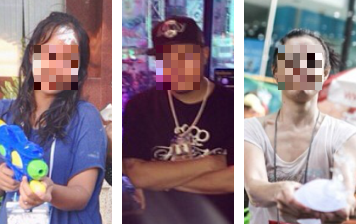} &
  \includegraphics[width=0.15\linewidth]{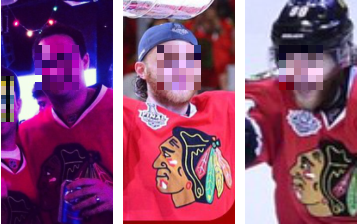} &
  \includegraphics[width=0.15\linewidth]{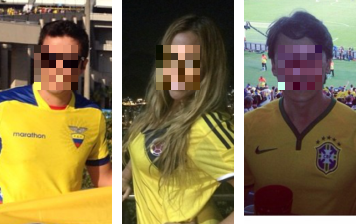} \\
  \textbf{City} & Bangkok & Seattle & Bangkok & Chicago & Rio \\
  \textbf{Attribute} & Yellow color & No sleeves & T-shirt & Red color & Yellow color \\
  \textbf{Month} & 2014 Dec, 2015 Dec & 2014 Oct & 2014 Apr & 2014 Jun & 2014 Jun, 2014 Jul \\
  \textbf{Keywords} & dad, father & halloween, freaknight & songkran, festival & cup, stanleycup & worldcup, brasil \\
  \bottomrule
\end{tabular}

\caption{Top five events detected across the world by finding anomalous behaviour in trends using methods from Section \ref{ssec:trend_prediction}. The words from the captions of the image posts are sorted by their TF-IDF scores in the associated event week (top-2 are shown). Images from each event are sorted based on number of terms in their caption matching the top-5 keywords.}
  \label{tab:events}
\end{table*}

\smallskip\noindent\textbf{Quantitative evaluation:}
Quantitative evaluation of our discovered events is challenging because there is no dataset or annotations of all the significant social events in the world.
However, we can check if the events we discover do in fact correspond to real social events, which can be construed as a kind of precision.

To do this evaluation, we manually inspect each discovered event and the associated top keywords to see if they reveal an understandable explanation: a real social event.
We measure the percentage of events with saliency greater than a threshold for which we found such a reason. 
Figure \ref{fig:captions} shows this percentage as a function of the saliency threshold.  
We find that \textbf{100\%} of the most salient events and \textbf{60\%} of all events have explainable reasons, indicating both the ability of our model to detect events and its ability to identify appropriate keywords for them.
Not surprisingly, the percentage of explainable events decreases as event saliency decreases, which validates our model's estimate of saliency as a measure of probability of corresponding to a real-world explainable event.
\begin{figure}
\centering
\includegraphics[width=0.49\linewidth]{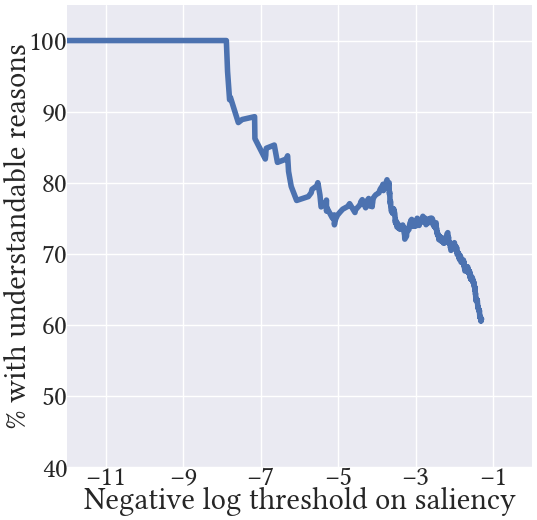}
\includegraphics[width=0.49\linewidth]{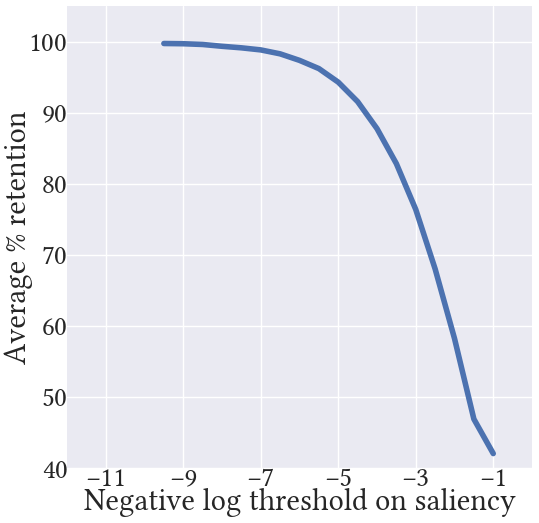}
\caption{\textbf{Left:} The percentage of events with saliency greater than a threshold that are explainable, plotted as the threshold varies.
	\textbf{Right:} The percentage of events retained when another sample with replacement is used for detection. 
}
\label{fig:captions}
\end{figure}

We also evaluate the robustness of our event detector by measuring the stability of detected events across random subsets of data.
We resample the dataset 20 times with replacement, and run both the trend characterization and event detection on each subset.
We then measure the fraction of outliers with saliency greater than a threshold in one sampled set that are still salient in a second set.
We call this fraction the \emph{retention}, and plot it in Figure \ref{fig:captions}.
Ideally, we want all salient events we detect in one dataset to be detected in all datasets, yielding high retention.
Indeed, the high saliency events are retained in other folds.
Furthermore, this retention rate increases consistently as the threshold value on saliency increases, indicating that the reciprocal of $p$-value is indeed a good measure of the saliency of events.

\subsection{Style trend analysis}\label{ssec:style_res}
Finally, we run the same trend analysis and event detection pipeline on style clusters.
Table \ref{tab:prediction} shows the prediction error of our parametric trend analysis compared to various baselines when making long-term fine-grained predictions over the next 26 weeks.
We find that our approach again significantly outperforms all baselines, and by a larger margin.

Figure \ref{fig:fig1} shows the most salient style-based events for selected cities. 
We find that with style clusters, we are able to identify events that involve attribute combinations, e.g., people wearing glasses with sleeveless tops during the ACL festival in Austin.
More striking are events such as Durga Puja in Kolkata or Fashion Week in Mumbai which are discovered in spite of the fairly nuanced associated appearance.

\subsection{Cross-dataset generalization}\label{ssec:domain_res}
We also show that our method generalizes well to cities not seen during CNN training.
We collected Flickr images from Barcelona (a city not in~\cite{matzen-17})
from 2013 to mid-2018 and fed them through the pipeline described in Section \ref{sec:method}. 
We detected a total of 97k people in these photos.

We test the predictability of our trend prediction method
on this unseen set of images. 
We used images from 2013 to mid-2017 to fit trends, then predicted the trend for the final year of data. 
Our model (MAE$=$0.043) performs significantly better than the best baseline, Autoregression (MAE$=$0.047), 
although fitting a sinusoid with a linear component also gives comparable performance (MAE$=$0.043).
We suspect this is because Barcelona does not see significant variations in weather~\cite{barca} and hence a smoother sinusoid models the seasonal changes as well as our model.

We also discovered events in Barcelona using the method described in Section \ref{ssec:events}.
The top-most event discovered in Barcelona corresponds to people gathering in yellow shirts for the ``\textbf{Catalan Way}'', a long human chain in support of Catalan independence from Spain, in September 2013 (Figure \ref{fig:topimg_barcelona}).
This event is a significant political event, and it validates our framework's ability to identify important social events from raw data across multiple datasets and bring them to the fore.

\begin{figure}
\centering
\includegraphics[width=0.8\linewidth]{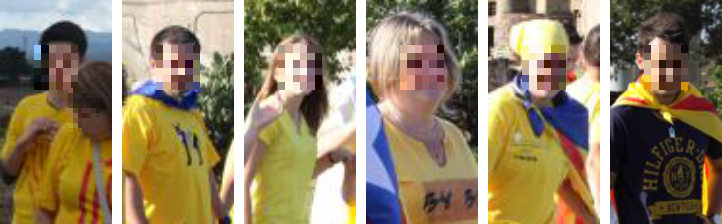}
\caption{Images from ``Catalan Way'' an event discovered from September 2013 in Barcelona.}
\label{fig:topimg_barcelona}
\end{figure}

\section{Conclusion and Future Work}
This work has established a framework for automatically analyzing temporal trends in fashion attributes and style by examining millions of photos published publicly to the web.
We characterized these trends using a new model that is both more interpretable and makes better long-term forecasts.
We also presented a methodology to automatically discover social events that impact how people dress.
However, this is but a first step and there are many questions still to be answered, such as the identification and mitigation of biases in social media imagery, and the propagation of styles across space.
The problem of analyzing trends is also relevant in other visual domains, such as understanding which animals are getting rarer over time in camera trap images~\cite{beery-18} or how land-use patterns are changing in satellite imagery~\cite{jean-19}.
We therefore believe that this is an important problem deserving of future research.

\smallskip\noindent\textbf{Acknowledgements.} This work was funded by NSF (CHS: 1617861 and CHS: 1513967) and an Amazon Research Award.

{\small
\bibliographystyle{ieee_fullname}
\bibliography{streetstyle-bibliography}
}

\end{document}